\documentclass[runningheads]{llncs}

\usepackage{eccv}

\usepackage{eccvabbrv}

\usepackage{graphicx}
\usepackage{booktabs}
\usepackage{algorithm}
\usepackage{algpseudocode}

\usepackage[accsupp]{axessibility}  

\usepackage[pagebackref,breaklinks,colorlinks]{hyperref}

\usepackage{orcidlink}

\begin{document}

\title{Adversarial Defense Teacher for Cross-Domain Object Detection under Poor Visibility Conditions} 


\author{Kaiwen Wang\inst{1} \and
Yinzhe Shen\inst{1} \and
Martin Lauer\inst{1}}

\authorrunning{K. Wang et al.}

\institute{Karlsruhe Institute of Technology (KIT)}

\maketitle

\begin{abstract}
  Existing object detectors encounter challenges in handling domain shifts between training and real-world data, particularly under poor visibility conditions like fog and night. Cutting-edge cross-domain object detection methods use teacher-student frameworks and compel teacher and student models to produce consistent predictions under weak and strong augmentations, respectively. In this paper, we reveal that manually crafted augmentations are insufficient for optimal teaching and present a simple yet effective framework named \textit{\textbf{A}dversarial \textbf{D}efense \textbf{T}eacher} (\textbf{ADT}), leveraging adversarial defense to enhance teaching quality. Specifically, we employ adversarial attacks, encouraging the model to generalize on subtly perturbed inputs that effectively deceive the model. To address small objects under poor visibility conditions, we propose a Zoom-in Zoom-out strategy, which zooms-in images for better pseudo-labels and zooms-out images and pseudo-labels to learn refined features. Our results demonstrate that ADT achieves superior performance, reaching 54.5\% mAP on Foggy Cityscapes, surpassing the previous state-of-the-art by 2.6\% mAP. Our codes will be available at https://anonymous.4open.science/r/ADT-3DC1/.

\end{abstract}

\section{Introduction}
\label{sec:intro}
As computer vision techniques progress, there has been a notable breakthrough in the performance of object detection \cite{fasterrcnn,yolo,cascadercnn,fcos}. A key driving force behind this advancement is the availability of large-scale annotated training data. The majority of large-scale image datasets \cite{kitti,argoverse,apollo,cityscapes,bddv} employed to train computer vision algorithms are recorded under normal visual conditions, such as daytime and clear weather. However, a noticeable decline in performance occurs when these algorithms are exposed to poor visibility conditions, \eg, fog, night, \etc \cite{adverse}.

To overcome this performance drop, supervised methods require extensive annotations, which is expensive and impractical. Cross-domain Object Detection (CDOD) \cite{dafasterrcnn,sw,umt,pt,at,cmt} has thus been proposed to address this issue, where a pre-trained object detector is adapted from a labeled source domain to an unlabeled target domain. As a semi-supervised learning scheme, CDOD eliminates the need for annotating training data in the target domain, making it more practical for real-world applications. 

\begin{figure}[tb]
  \centering
  \includegraphics[width=\textwidth]{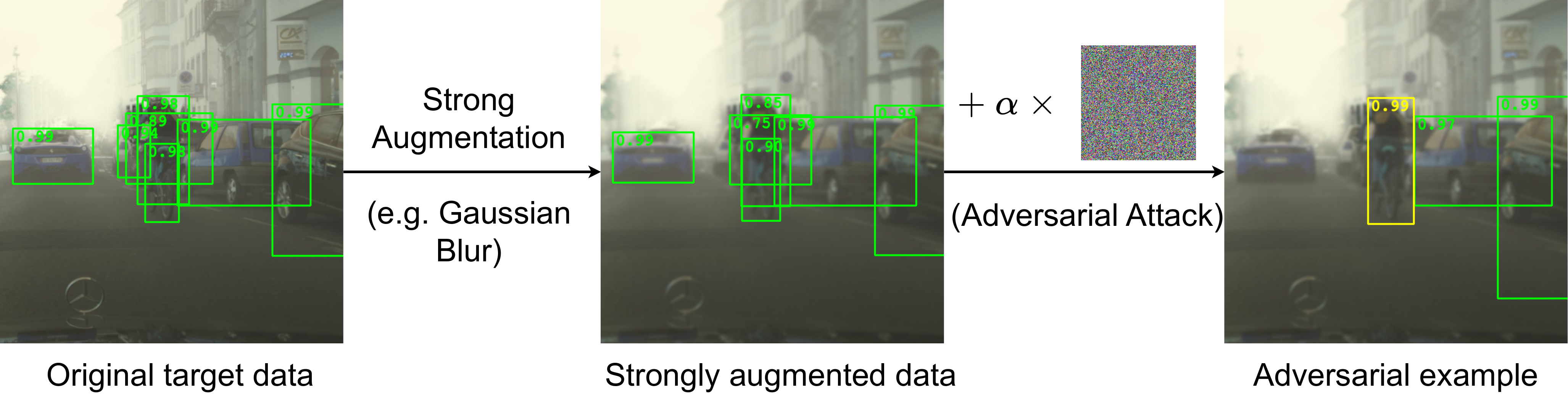}
  \caption{Current self-training methods enforce consistent predictions on original and strongly augmented data. However, this can be insufficient, as manual augmentation visibly alters the appearance while the model maintains similar predictions as on the original data. In contrast, we add an additional adversarial perturbation to the augmented data, which remains imperceptible to humans (thus within the same domain), but effectively deceives the model. The adversarial attack induces highly inconsistent predictions, thereby improving the mutual learning quality. Green boxes denote true positives while yellow boxes indicate misclassifications. Best viewed in color.}
  \label{fig:fig1}
\end{figure}

The most straightforward category of CDOD involves aligning two domains through image-level or instance-level adversarial learning \cite{dafasterrcnn,sw}, aiming to acquire a domain-invariant representation. While this approach has proven effective, relying solely on adversarial learning for complex recognition tasks such as object detection still results in a significant performance gap compared to the oracle model (fully supervised on the target domain). In recent years, the self-training paradigm \cite{umt,pt,at,cmt} has shown more promising results in mitigating domain shift by using teacher-student mutual learning \cite{mt}. Specifically, the entire model consists of two architecturally identical components: a student model and a teacher model. The student model is trained through standard gradient updating, while the teacher model is updated using the exponential moving average (EMA) of the weights from the student model. Additionally, the consistency loss between the pseudo-labels predicted by the teacher model on weakly augmented data and the predictions of the student model on strongly augmented data guides the adaptation mutually. Prior methods \cite{at,pt,ht,cmt} suggest employing manually designed augmentations, \eg, Gaussian blur and grayscaling, to deceive the model. However, one major challenge lies in the low teaching quality due to the low effectiveness of manually crafted data augmentation. As shown in \cref{fig:fig1}, the predictions on original target data and strongly augmented data exhibit notable similarity. The slight inconsistency in predictions diminishes the overall quality of the teacher-student mutual learning process. 

To address this issue, we propose a simple yet effective framework called Adversarial Defense Teacher (ADT), which employs adversarial defense to enhance the quality of teacher-student mutual learning. As presented in \cite{fgsm,pgd}, introducing an imperceptible adversarial perturbation to the input, known as an adversarial attack, can lead to the misclassification or deception of neural networks. The process of training models to resist and handle adversarial examples is referred to as adversarial defense. Leveraging adversarial defense for teacher-student frameworks has three advantages. Firstly, the introduction of adversarial perturbations induces highly inconsistent predictions in terms of both regression and classification, thereby enhancing the effectiveness of mutual learning. After conducting an additional adversarial attack on the augmented data in \cref{fig:fig1}, the output of the same network changes notably. The rider is misclassified as a pedestrian with a high confidence of 0.99, and several objects are no longer recognized as objects. Meanwhile, the regression quality of the rightmost vehicle experiences a significant drop. Secondly, these adversarial perturbations, while impactful on model predictions, remain imperceptible to humans, ensuring that the resulting adversarial example continues to belong to the same domain. Hence, the robustness of student models is enhanced. Thirdly, the teacher model, constructed as the EMA of student models, attains increased robustness, consequently producing pseudo-labels of higher quality.

While adversarial defense significantly increases the robustness to detect objects, the detection of small and obscure objects remains a challenge under adverse visibility conditions. To tackle this issue, we introduce a Zoom-in Zoom-out strategy. Target images are zoomed in before feeding into the teacher model so that smaller objects are upscaled and thus more likely to be included in the pseudo-labels. Subsequently, we perform a zoom-out operation on both the image and pseudo-labels with the same ratio. The student model is then forced to detect downscaled objects, ensuring that it benefits from extracting finer features.

Equipped with these perspectives, we make the following contributions.
\begin{itemize}
    \item We introduce Adversarial Defense Teacher (ADT), a teacher-student framework that uses adversarial defense to enhance the quality of mutual learning. This approach encourages model generalization on subtly perturbed inputs that effectively deceive the model, thereby ensuring a more robust performance on the target domain. 
    \item To tackle the issue of challenging small object detection under adverse conditions, we introduce a Zoom-in Zoom-out strategy, allowing the teacher model to improve the recall of pseudo-labels during the zoom-in phase and refining features for the student model through a zoom-out operation.
    \item We conduct extensive experiments and verify the effectiveness of our framework, which outperforms all existing SOTA by a large margin. Our approach achieves 54.5\% mAP on Foggy Cityscapes, surpassing the previous SOTA by 2.6\%.
\end{itemize} 
\section{Related Works}
\label{sec:related}
\subsection{Object Detection under Poor Visibility Conditions}
Object detection aims at classifying and localizing the objects given in an input image. Depending on whether region proposal and object classification are conducted simultaneously or sequentially, the architecture of object detectors can be primarily divided into two categories: single-stage \cite{yolo,yolo9000,yolov3,retinanet,fcos} and two-stage\cite{fasterrcnn,cascadercnn}. Single-stage detectors such as YOLO series \cite{yolo,yolo9000,yolov3} perform object localization and classification in a single forward pass through the neural network. SSD \cite{ssd} effectively manages objects of diverse scales by incorporating multiple bounding boxes with different aspect ratios at each spatial location in the feature map. On the other hand, two-stage methods first propose regions of interest (ROIs) by utilizing region proposal networks (RPN) \cite{fasterrcnn} and refine those candidates in the second stage. In this work, cross-domain object detection is explored using Faster R-CNN \cite{fasterrcnn} as the baseline architecture considering its wide range of applications.

Despite the great success, object detection under poor visibility conditions, like fog, rain and night, has been proven to be vulnerable \cite{adverse}. Several approaches have been proposed to address this challenge. In \cite{multi_modal2}, a multimodal sensor information technique is proposed to improve object detection under adverse weather. In \cite{haze_removal1,haze_removal2,rainstreak_removal,raindrop_removal}, input images are preprocessed to enhance the visibility conditions, such as removing haze, rain streaks and raindrops. Despite these advancements, a substantial gap persists in achieving precise object detection under challenging visibility conditions.
    \subsection{Cross-Domain Object Detection}
Recently, considerable works employ domain adaptation to achieve better object detection under challenging visibility conditions. Such cross-domain object detection approaches can be mainly divided into three categories: feature alignment, domain translation and self-training. Feature alignment methods \cite{dafasterrcnn,sw,multi,maf,selective,htcn,crda} conduct adversarial learning to align the features from both domains with a gradient reverse layer (GRL). Domain translation methods \cite{translation1,translation2,translation3} aim at translating the source data into target-like styles and thus improve the performance of CDOD. Recently, self-training methods \cite{umt,pt,at,cmt,ht} use weak-strong augmentation and Mean Teacher (MT)\cite{mt} for teacher-student mutual learning and have demonstrated superior advantages in this field. Adaptive Teacher (AT) \cite{at} additionally applies adversarial learning to bridge the domain gap. Probabilistic Teacher (PT) \cite{pt} and Harmonious Teacher (HT) \cite{ht} focus on improving the quality of pseudo-labels for both classification and regression. Contrastive Mean Teacher (CMT) \cite{cmt} leverages contrastive learning to optimize object-level features.
\subsection{Adversarial Attack and Adversarial Defense}
Different from adversarial learning \cite{at} which focuses on learning domain-invariant features, adversarial defense aims at enhancing the robustness of neural networks against intentional yet imperceptible perturbations, \ie, adversarial attacks. Adversarial attacks are broadly categorized into two types: white-box and black-box attacks. In a white-box attack, the adversary has complete knowledge of the network under attack, including its architecture, parameters and training data. Fast Gradient Sign Method (FGSM) is proposed in \cite{fgsm} to leverage the gradients of a neural network to design adversarial examples. Building upon this, Projected Gradient Descent \cite{pgd} employs multi-step perturbations for a more powerful attack. In contrast, a black-box attacker only has limited or no knowledge about the system, making the attack more challenging due to the absence of internal details.
\section{Adversarial Defense Teacher}
\label{sec:adt}

\begin{figure}[tb]
  \centering
  \includegraphics[width=\textwidth]{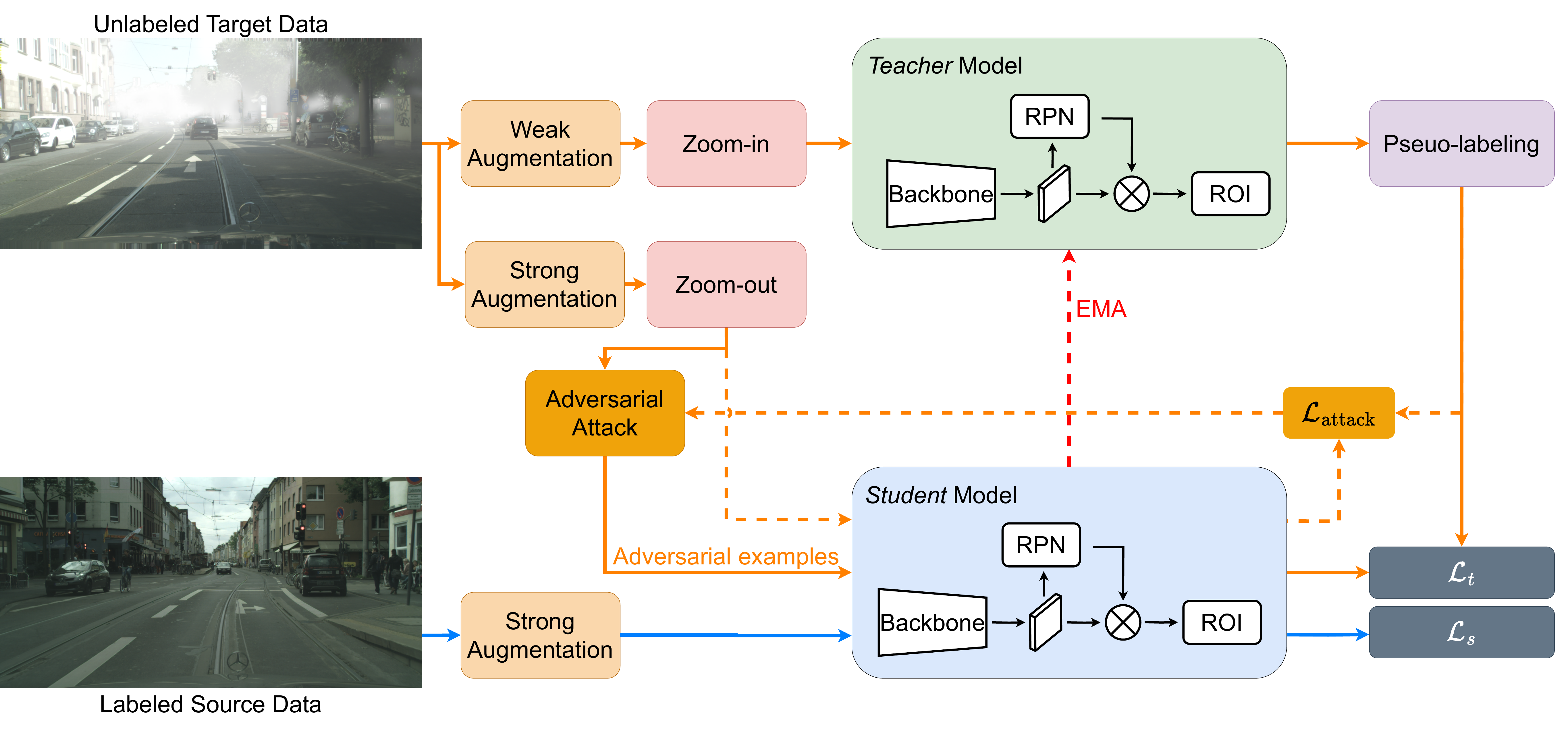}
  \caption{\textbf{Overview of the proposed Adversarial Defense Teacher.} Our model includes two branches: 1) supervised branch (blue lines): strongly augmented source data is fed into the student model. 2) unsupervised branch (orange lines): the teacher model processes weakly augmented and zoomed-in data to generate pseudo-labels with high confidence. Adversarial attacks (dashed lines) are conducted on the student model based on the inconsistency loss $\mathcal{L}_{\text{attack}}$ between pseudo-labels and predictions on strongly augmented and zoomed-out data. The resulting adversarial examples are reintroduced to the student model. Best viewed in color.}
  \label{fig:adt}
\end{figure}

In this section, we introduce the proposed Adversarial Defense Teacher (ADT) framework in \cref{fig:adt}. We initially describe the weak-strong augmentation-based MT paradigm for CDOD briefly in Sec. \ref{ssec:adt:mt} and analyze existing issues in Sec. \ref{ssec:adt:analysis}. Subsequently, we delve into the design of our ADT in Sec. \ref{ssec:adt:ad} and \ref{ssec:adt:resize}.
\subsection{Weak-Strong Augmentation-Based Mean Teacher}
\label{ssec:adt:mt}
CDOD aims at mitigating the impact of domain shift between the source domain  $\mathcal{D}_s=\{X_s,B_s,C_s\}$ and the target domain $\mathcal{D}_t=\{X_t\}$ on object detectors. Source images $X_s$ are labeled with corresponding bounding box annotations $B_s$ and class labels $C_s$, while target images $X_t$ are not annotated. In the context of poor visibility conditions, the source domain denotes clear weather in the daytime, while the target domain encompasses various poor visibility conditions, such as fog and night.

State-of-the-art methods \cite{pt,at,cmt,ht} employ the MT framework \cite{mt} and weak-strong augmentation for CDOD. A source model is first pre-trained on labeled source data, serving as the initial model for two architecturally identical models: the teacher and the student model. The teacher model generates pseudo-labels with high confidence $B_t'$ and $C_t'$ on weakly augmented (\eg, random horizontal flip, \etc) target samples $X_t^w$, while the student model is trained on both the labeled source data $\{X_s,B_s,C_s\}$ and strongly augmented (\eg, Gaussian blur, grayscaling, \etc) target data $\{X_t^s,B_t',C_t'\}$. The consistency loss between the pseudo-labels generated by the teacher model on weakly augmented target samples $X_t^w$ and the predictions of the student model on strongly augmented target samples $X_t^s$ improves the generalization capability of the student model through gradient back-propagation. Concurrently, the teacher model is updated through the EMA of the weights of the student model, performed without any gradient involvement.

Taking Faster R-CNN \cite{fasterrcnn} as an example, the optimization objective of the student model on labeled source domain and pseudo-labeled target domain can be respectively written as follows:
\begin{equation}
    \begin{aligned}
      \mathcal{L}_{s}(X_s,B_s,C_s) = \mathcal{L}^{\text{rpn}}(X_s,B_s,C_s)+\mathcal{L}^{\text{roi}}(X_s,B_s,C_s)
    \end{aligned}
    \label{eq:ls}
\end{equation}

and 
\begin{equation}
    \begin{aligned}
      \mathcal{L}_{t}(X_t^s,B_t',C_t') = \mathcal{L}^{\text{rpn}}(X_t^s,B_t',C_t')+\mathcal{L}^{\text{roi}}(X_t^s,B_t',C_t')
    \end{aligned}
    \label{eq:lt}
\end{equation}
where $\mathcal{L}^{\text{rpn}}$ indicates the loss of Region Proposal Network (RPN) and $\mathcal{L}^{\text{roi}}$ is the loss of Region of Interest (ROI). Both losses include regression and classification branches.

\begin{equation}
    \begin{aligned}
      \mathcal{L}^{\text{rpn}}=\sum_i \mathcal{L}^{\text{rpn}}_{\text{cls}}(p_i,p_i^*)+\sum_{p_i^*=1}\mathcal{L}^{\text{rpn}}_{\text{reg}}(t_i,t_i^*)
    \end{aligned}
    \label{eq:rpn}
\end{equation}
Here, $i$ denotes the index of an anchor and $p_i$ represents the predicted probability of anchor $i$ being an object. Its ground truth $p_i^*$ is 1 if the anchor is positive and 0 if the anchor is negative. $t_i$ and $t_i^*$ are respectively the prediction and ground truth of regression offsets. Background is ignored for $\mathcal{L}_{\text{reg}}$.

\begin{equation}
    \begin{aligned}
      \mathcal{L}^{\text{roi}}=\sum_j \mathcal{L}^{\text{roi}}_{\text{cls}}(q_j,q^*_j)+\sum_{q^*_j\geq1}\mathcal{L}^{\text{roi}}_{\text{reg}}(o_j,o_j^*)
    \end{aligned}
    \label{eq:roi}
\end{equation}
Similarly, $j$ is the index of a region of interest (ROI) and $q_j$ is a vector indicating the predicted classification probability. The ground truth of ROIs matched to the background is labeled $q^*_j=0$, and these ROIs are not considered for $\mathcal{L}_{\text{reg}}$.

The evolving student model creates a more robust teacher model in turn. The weights of the teacher model are updated as follows:
\begin{equation}
    \theta_{\text{teacher}} \leftarrow \beta \theta_{\text{teacher}} + (1-\beta)\theta_{\text{student}}
    \label{eq:ema}
\end{equation}
where $\theta_{\text{teacher}}$ and $\theta_{\text{student}}$ denote the weights of the teacher and student model respectively.

\subsection{Analysis on Weak-Strong Augmentation-Based Mean Teacher}
\label{ssec:adt:analysis}

\begin{figure}[tb]
  \centering
  \includegraphics[width=0.7\textwidth]{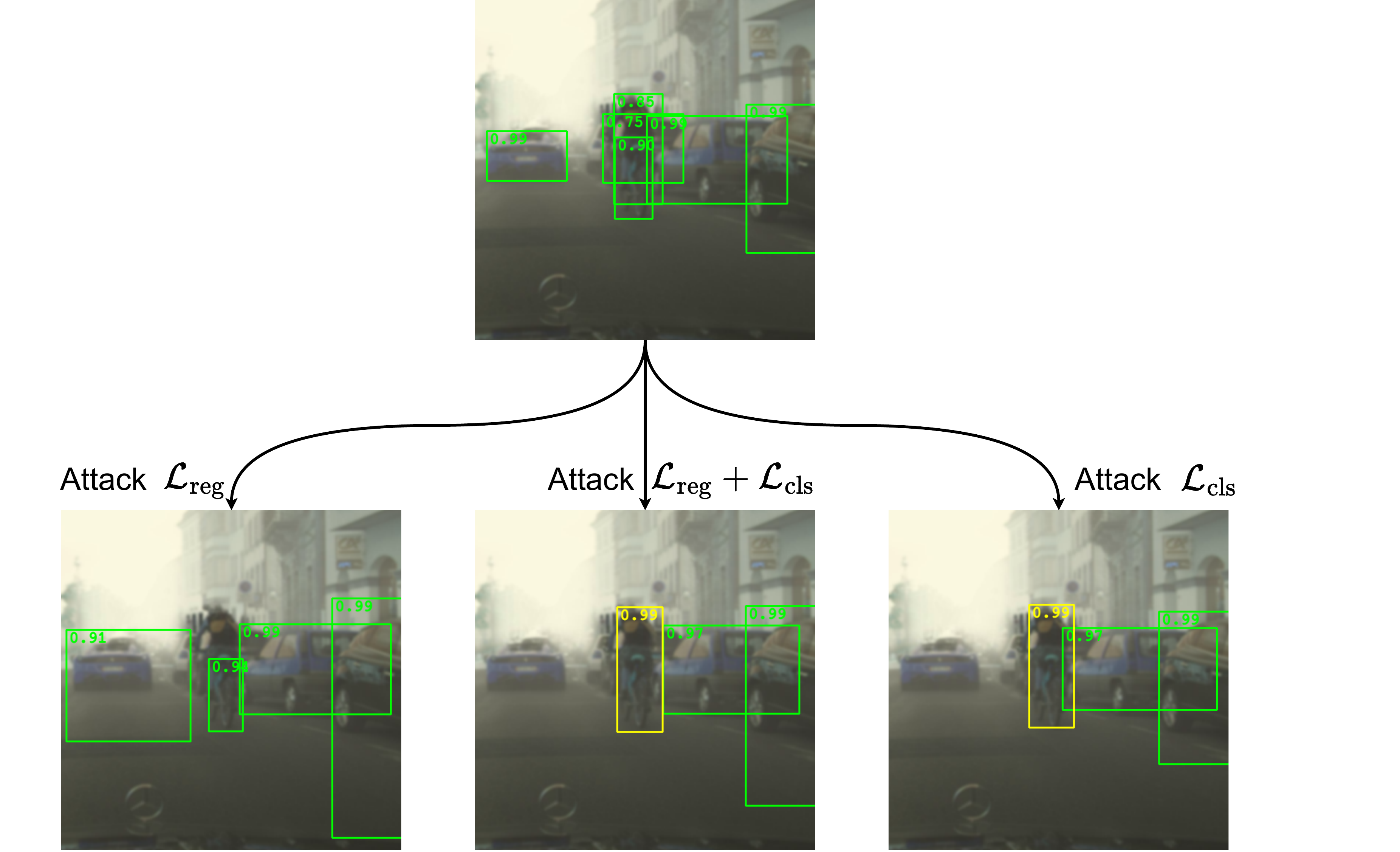}
  \caption{Conducting adversarial attacks based on various losses leads to different deceptions. Green boxes denote true positives while yellow boxes indicate misclassifications. Best viewed in color.}
  \label{fig:analysis}
\end{figure}

As described in \cref{ssec:adt:mt}, the mutual learning process between the teacher and student model comprises two key components: the EMA process and the teaching process. The EMA process is equivalent to an implicit model ensemble and thus provides more robust pseudo-labels. Meanwhile, the teaching process is dedicated to refining the performance of the student model on the target domain, guided by the pseudo-labels generated by the teacher model. 

Existing approaches \cite{pt,at,cmt,ht} leverage different augmentation strategies on teacher and student models. The teacher model generates reliable pseudo-labels on weakly augmented data, while the student model is trained to produce consistent predictions on strongly augmented data. However, one drawback emerges. The randomly selected strong augmentation may not guide the student model in the most informative direction, \ie, the direction where the student model makes the most inconsistent predictions. As shown in \cref{fig:fig1}, the predictions on blurred data closely resemble those on the original target data, making the teacher-student mutual learning less effective. An optimal strong data augmentation should strike a balance: avoiding significant modifications to the visual appearance to the extent that the augmented data no longer belongs to the same domain, while still introducing sufficient perturbations to impact predictions meaningfully. Conducting a human-imperceptible adversarial attack fulfills this requirement perfectly.

We further illustrate this problem in \cref{fig:analysis}, where we take a sample from the Foggy Cityscapes dataset as an example and conduct adversarial attacks. Object detection loss includes regression loss and classification loss (see \cref{eq:ls}). When deceiving the model by making slight modifications to the input image to solely increase the regression loss $\mathcal{L}_{\text{reg}}$, most objects can still be recognized. Although the objects don't seem to be occluded to humans, the boundaries predicted by the model are no longer as precise as they would be without adversarial attacks. Meanwhile, if we attack the model to specifically increase the classification loss $\mathcal{L}_{\text{cls}}$, misclassifications may occur while maintaining high regression quality. Adversarial attacks based on the entire detection loss (\ie, $\mathcal{L}_{\text{reg}}+\mathcal{L}_{\text{cls}}$) combine both effects, yielding challenging examples where the model produces the most inconsistent predictions. Adversarial defense on such challenging examples effectively improves the mutual learning quality and overall performance as well.

\subsection{Adversarial Defense for CDOD}
\label{ssec:adt:ad}
In this section, we first briefly introduce two white-box adversarial defense methods for image classification, including Fast Gradient Sign Method (FGSM) \cite{fgsm} and Projected Gradient Descent (PGD) \cite{pgd}, and then describe how we apply PGD to enhance CDOD.

\subsubsection {Fast Gradient Sign Method.} FGSM stands as a widely employed adversarial attack method in image classification. Its mechanism involves leveraging the gradients of a neural network to craft an adversarial example. Given an input image, the method uses the gradients of the loss concerning the input image to generate a new image that maximizes the loss. This generated adversarial example can be expressed mathematically as in \cref{eq:fgsm}.
\begin{equation}
    x_{\text{adv}}=x+\alpha\cdot \text{sgn}(\nabla_x \mathcal{L}(\theta,x,y))
    \label{eq:fgsm}
\end{equation}
where the parameter $\alpha$ governs the magnitude of the adversarial perturbation, and $\text{sgn}(\cdot)$ refers to the sign function. Furthermore, $\mathcal{L}(\theta,x,y)$ indicates the loss between the prediction and the ground truth $y$. FGSM has been proven to be a fast one-step method for adversarial attacks on neural networks. However, its effectiveness is surpassed by multi-step variants, such as Projected Gradient Descent (PGD).

\subsubsection{Projected Gradient Descent.} PGD is a more powerful adversarial attack method that iteratively applies FGSM with additional constraints to generate adversarial examples as in \cref{eq:pgd}.
\begin{equation}
    x_{\text{adv}}^{t}=\left\{\begin{aligned}
    & x\ & t=0\\ & \text{Clip}_{x,\epsilon}\left(x_{\text{adv}}^{t-1}+\alpha\cdot \text{sgn}(\nabla_x \mathcal{L}(\theta,x_{\text{adv}}^{t-1},y))\right) &\text{otherwise}
    \end{aligned}\right.
    \label{eq:pgd}
\end{equation} 
where $x_{\text{adv}}^{t}$ denotes the adversarial example at iteration $t$, and $\text{Clip}_{x,\epsilon}(\cdot)$ ensures that the adversarial example remains within the $\epsilon-$neighborhood around the original image. The iterative process aims to find the perturbation that maximizes the loss function while staying within a certain bounded region.

\subsubsection{Apply PGD for CDOD.} White-box adversarial attacks such as FGSM and PGD require the attacker to have complete knowledge of the model's architecture, parameters and training data so that the gradient can be computed and the vulnerability of the model can be exploited more efficiently. In the Mean Teacher mutual learning paradigm, the access to both the teacher and student models, along with their parameters, aligns with the requirements of white-box attacks. However, white-box attacks also require ground truth to calculate the gradient, which is not available for the unlabeled target data. To achieve adversarial attack and defense under the setting of CDOD, we take the pseudo-labels generated by the teacher model as ground truth and calculate an attack loss $\mathcal{L}_{\text{attack}}$ for the student model. Based on adversarial examples generated to increase $\mathcal{L}_{\text{attack}}$, the student model is trained to defend itself from such attacks. With each student model being more robust, the teacher model is expected to benefit as well.

Consistent with common CDOD methods \cite{umt,at}, we also employ a confidence threshold to filter pseudo-labels. The pseudo-labels typically exhibit a higher precision than recall. This implies that positive objects identified in pseudo-labels are likely correct, yet there is a notable risk of overlooking a portion of actual positive instances, \ie, false negatives. Consequently, in the process of conducting adversarial attacks, our objective is to deceive the model in a way that disrupts its accurate detection of positive objects within pseudo-labels. It is crucial, however, to ensure that the adversarial attack has minimal to no impact on regions recognized by the teacher model as background, as these areas may contain false negatives. Following this concept, we redesigned RPN and ROI loss as follows instead of using \cref{eq:rpn} and \cref{eq:roi}.

\begin{equation}
    \begin{aligned}
      \mathcal{L}^{\text{rpn}}_{\text{attack}}=\sum_{p_i^*=1} \mathcal{L}^{\text{rpn}}_{\text{cls}}(p_i,p_i^*)+\sum_{p_i^*=1}\mathcal{L}^{\text{rpn}}_{\text{reg}}(t_i,t_i^*)
    \end{aligned}
    \label{eq:rpn_attack}
\end{equation}

\begin{equation}
    \begin{aligned}
      \mathcal{L}^{\text{roi}}_{\text{attack}}=\sum_{q^*_j\geq1} \mathcal{L}^{\text{roi}}_{\text{cls}}(q_j,q^*_j)+\sum_{q^*_j\geq1}\mathcal{L}^{\text{roi}}_{\text{reg}}(o_j,o_j^*)
    \end{aligned}
    \label{eq:roi_attack}
\end{equation}

As shown in \cref{fig:adt}, we conduct adversarial attacks on the strongly augmented data, making it more challenging for the model to produce consistent predictions. The overall loss used for adversarial attacks is defined as:
\begin{equation}
    \begin{aligned}
      \mathcal{L}_{\text{attack}}=\mathcal{L}^{\text{rpn}}_{\text{attack}}(X_t^s,B_t',C_t')+\mathcal{L}^{\text{roi}}_{\text{attack}}(X_t^s,B_t',C_t')
    \end{aligned}
    \label{eq:attack}
\end{equation}

After applying PGD on the student model using $\mathcal{L}_{\text{attack}}$, adversarial examples are generated and then fed into the student model again. Due to the adversarial attack, the student model is deceived effectively to output highly inconsistent predictions as the pseudo-labels predicted by the teacher model, which in turn enhances the effectiveness of teacher-student mutual learning. 

\subsection{Zoom-in Zoom-out Strategy}
\label{ssec:adt:resize}
After conducting adversarial defense, the model is robust enough to correctly detect objects despite various augmentations or perturbations. However, particularly under challenging visibility conditions, obscure objects of smaller sizes are still hard to capture. This challenge arises from the low recall of pseudo-labels, hindering the model's ability to effectively identify small objects. When encountering obscure objects, humans often exhibit a natural tendency to zoom in on an image for closer inspection. This zoom-in behavior facilitates the recognition of obscured details, allowing for improved comprehension of the object. Remarkably, even after zooming out again, humans can retain the ability to recognize previously obscure objects. Inspired by this, we propose a Zoom-in Zoom-out strategy, intending to increase the recall of pseudo-labels predicted by the teacher model and encourage the student model to extract detailed features.

During the zoom-in phase, the teacher model takes the zoomed-in image as input so that smaller objects can be better detected, thereby increasing the recall of the generated pseudo-labels. On the other hand, we zoom out the image and corresponding pseudo-labels before feeding them into the student model. As all objects are downscaled, the student model is enforced to extract features from smaller details. To prevent the size distribution from being disturbed by this strategy, zoomed-out pseudo-labels that are smaller than a specific threshold are removed.
\section{Experiments}
\label{sec:exp}
\subsection{Datasets}
\label{ssec:exp:datasets}
To validate the effectiveness of our approach, we conduct experiments on multiple benchmarks, including 1) fog: adaptation from normal weather to foggy weather, 2) night: adaptation from daytime to night. The public datasets used in our experiments are as follows:
\paragraph{Cityscapes.} Cityscapes \cite{cityscapes} contains 2975 training images and 500 validation images, all captured in normal weather conditions. Each image is annotated with pixel-level labels. We convert the instance segmentation labels into bounding box annotations for our experiments.
\paragraph{Foggy Cityscapes.} Foggy Cityscapes \cite{foggy_cityscapes} is generated by adding synthesized fog on images in the Cityscapes. Each image is rendered with three levels of fog (0.005, 0.01, 0.02), representing the visibility ranges of 600, 300 and 150 meters.
\paragraph{BDD100K.} BDD100K \cite{bdd} is a large-scale driving dataset consisting of 100k images that include various visibility conditions, such as daytime and night. Following \cite{2pc}, we employ the daytime subset as the source domain and the night subset as the target domain. This results in 36728 source training images, 17961 target training images and 3929 target validation images.

\subsection{Implementation Details}
\label{ssec:exp:detail}
Following prior studies such as \cite{dafasterrcnn,sw,umt,pt,at,cmt}, our Adversarial Defense Teacher (ADT) relies on Faster R-CNN \cite{fasterrcnn} as the foundational detection model, implemented using the Detectron2 framework \cite{detectron2}. To underscore the adaptability of our approach, diverse backbones are employed. VGG 16 \cite{vgg} is chosen for fog adaptation, aligning with strategies from \cite{dafasterrcnn,sw,umt,pt,cmt,ht}, while ResNet-50 \cite{resnet} is employed for night adaptation following \cite{2pc}. These backbones are pre-trained on ImageNet \cite{imagenet}. In fog adaptation, we maintain consistency by resizing the shorter side of images to 600, preserving image ratios during both training and evaluation. A confidence threshold of $\delta=0.8$ is set for all experiments, and optimization is performed using Stochastic Gradient Descent (SGD). Data augmentation includes random horizontal flips for weak augmentation, and strong augmentations involve random color jittering, grayscaling, Gaussian blurring, and cut-out patches. The weight smoothing factor for the EMA ($\beta$ in \cref{eq:ema}) is set to 0.9996. Executed on 4 Nvidia GPU A100s, each experiment uses a batch size of 16 and is implemented in PyTorch. We report the average precision (AP) with a threshold of 0.5 for each class as well as the mean AP (mAP) over all classes for object detection following existing works \cite{dafasterrcnn,sw,umt,pt,at,ht,cmt} for all of the experimental settings.

\subsection{Fog: Cityscapes $\rightarrow$ Foggy Cityscapes}
\label{ssec:exp:fog}
Fog, a meteorological occurrence, occurs when water droplets or ice crystals gather in the air close to the ground, resulting in diminished visibility. Its impact on visibility, such as scattering and absorption of light, causes distant objects to appear blurry or obscured. In this experiment, we evaluate ADT on the commonly used benchmark Cityscapes $\rightarrow$ Foggy Cityscapes, where the object detector needs to overcome the domain shift from normal to foggy weather.
\begin{table}[tb]
\caption{\textbf{Results of Cityscapes $\rightarrow$ Foggy Cityscapes.} we employ VGG16 as the backbone for all fog adaptation experiments to ensure a fair comparison.}
\label{tab:c2f}
\resizebox{\textwidth}{!}{%
\begin{tabular}{@{}l|c|c|cccccccc|c@{}}
\toprule
Method                       & Reference & Split & person        & rider         & car           & truck         & bus           & train         & mcycle        & bicycle       & mAP           \\ \midrule
Source                       & -         & 0.02  & 25.9          & 29.4          & 35.4          & 6.9           & 19.8          & 4.3           & 16.1          & 22.7          & 20.1          \\
Oracle                       & -         & 0.02  & 41.9          & 48.1          & 64.3          & 29.1          & 52.0          & 38.7          & 35.7          & 42.5          & 44.0          \\ \midrule
DA-Faster\cite{dafasterrcnn} & CVPR'18   & 0.02  & 25.0          & 31.0          & 40.5          & 22.1          & 35.3          & 20.2          & 20.0          & 27.1          & 27.6          \\
SW\cite{sw}                  & CVPR'19   & 0.02  & 29.9          & 42.3          & 43.5          & 24.5          & 36.2          & 32.6          & 30.0          & 35.3          & 34.3          \\
UMT\cite{umt}                & CVPR'21   & 0.02  & 33.0          & 46.7          & 48.6          & 34.1          & 56.5          & 46.8          & 30.4          & 37.4          & 41.7          \\
PT\cite{pt}                  & ICML'22   & 0.02  & 40.2          & 48.8          & 59.7          & 30.7          & 51.8          & 30.6          & 35.4          & 44.5          & 42.7          \\
TDD\cite{tdd}                & CVPR'22   & 0.02  & 39.6          & 47.5          & 55.7          & 33.8          & 47.6          & 42.1          & 37.0          & 41.4          & 43.1          \\
AT\dag\cite{at}                  & CVPR'22   & 0.02  & 45.3          & 55.7          & 63.6          & 36.8          & 64.9          & 34.9          & 42.1          & \textbf{51.3} & 49.3          \\
CMT\cite{cmt}                & CVPR'23   & 0.02  & 45.9          & 55.7          & 63.7          & \textbf{39.6} & \textbf{66.0} & 38.8          & 41.4          & 51.2          & 50.3          \\
HT\cite{ht}                  & CVPR'23   & 0.02  & \textbf{52.1} & 55.8          & 67.5          & 32.7          & 55.9          & 49.1          & 40.1          & 50.3          & 50.4          \\ \midrule
ADT                          & Ours      & 0.02  & 49.4          & \textbf{57.9} & \textbf{67.6} & 35.8          & 55.4          & \textbf{51.9} & \textbf{42.2} & 48.6          & \textbf{51.1} \\ \midrule
Source                       & -         & All   & 35.1          & 37.8          & 51.4          & 16.6          & 22.8          & 12.6          & 26.4          & 36.5          & 29.9          \\
Oracle                       & -         & All   & 46.8          & 51.6          & 68.7          & 33.6          & 56.1          & 45.7          & 42.1          & 48.9          & 49.2          \\ \midrule
SW\ddag \cite{sw}                 & CVPR'19   & All   & 34.2          & 46.3          & 51.0          & 28.7          & 44.9          & 24.0          & 33.8          & 37.1          & 37.5          \\
PDA\cite{pda}                & WACV'20   & All   & 36.0          & 45.5          & 54.4          & 24.3          & 44.1          & 25.8          & 29.1          & 35.9          & 36.9          \\
ICR-CCR\cite{crda}           & CVPR'20   & All   & 32.9          & 43.8          & 49.2          & 27.2          & 36.4          & 36.4          & 30.3          & 34.6          & 37.4          \\
PT\cite{pt}                  & ICML'22   & All   & 43.2          & 52.4          & 63.4          & 33.4          & 56.6          & 37.8          & 41.3          & 48.7          & 47.1          \\
AT\cite{at}                  & CVPR'22   & All   & 45.5          & 55.1          & 64.2          & 35.0          & 56.3          & 54.3          & 38.5          & 51.9          & 50.9          \\
CMT\cite{cmt}                & CVPR'23   & All   & 47.0          & 55.7          & 64.5          & \textbf{39.4} & 63.2          & 51.9          & 40.3          & \textbf{53.1} & 51.9          \\ \midrule
ADT                          & Ours      & All   & \textbf{51.1} & \textbf{58.4} & \textbf{71.3} & 37.6          & \textbf{63.5} & \textbf{56.5} & \textbf{46.7} & 51.3          & \textbf{54.5} \\ \bottomrule
\end{tabular}%
}
\dag Results reproduced by CMT\cite{cmt}. \ddag Results reproduced by PT\cite{pt}.
\end{table}

The results are summarized in \cref{tab:c2f}. We present the CDOD training and evaluation results, comparing them with the performance of source (fully supervised on the source domain) and oracle models (fully supervised on the target domain). We consider both images with the highest fog severity ("0.02" split) and all images ("All" split) in Foggy Cityscapes. Similar to many other Mean Teacher-based approaches \cite{at,cmt,ht}, our method surpasses the performance of the oracle models on both splits, which are directly trained on the labeled target domain. This indicates the effectiveness of the teacher-student mutual learning framework in transferring cross-domain knowledge by leveraging images from both domains. 

Furthermore, our ADT outperforms all state-of-the-art approaches by a large amount (2.6\%). Among those methods, PT \cite{pt} and HT \cite{ht} enhance pseudo-label quality by combining classification and regression uncertainties. AT \cite{at} introduces adversarial learning to bridge the domain gap, while CMT \cite{cmt} leverages contrastive learning to optimize object-level features. However, these methods overlook the limited inconsistency between predictions on weakly augmented and strongly augmented target images, leading to suboptimal teacher-student mutual learning.

Notably, ADT surpasses the previous best performance (from CMT) by 0.8\% mAP on the "0.02" split and 2.6\% mAP on the "All" split. The relatively larger gain on the "All" split underscores ADT's robustness in learning from a more diverse set of unlabeled data. This is particularly crucial in real-world applications where acquiring abundant unlabeled data is feasible, but labeling them is resource-intensive. ADT's capacity to consistently improve target-domain performance aligns well with the evolving landscape of growing unlabeled data, making it well-suited for such real-world scenarios.

\subsection{Night: BDD100K Daytime $\rightarrow$ BDD100K Night}
\label{ssec:exp:night}
Poor visibility conditions at night are primarily due to the absence or reduction of natural light. During nighttime, the primary source of illumination is often artificial lighting, such as streetlights or vehicle headlights. The challenges in low-light or nighttime conditions include reduced contrast, limited visibility of dark objects, and the potential for glare from bright light sources. In this experiment, we evaluate ADT on the widely used BDD100K benchmark, focusing on the daytime-to-night domain shift scenario.The object detector is challenged to adapt and maintain effectiveness in transitioning from daylight to nighttime conditions.

\begin{table}[tb]
\caption{\textbf{Results of BDD100K daytime $\rightarrow$ BDD100K night.} To ensure a fair comparison, we employ ResNet50 as the backbone for all night adaptation experiments. The AP for the class "train" is not reported as it is 0 for all experiments.}
\label{tab:d2n}
\resizebox{\textwidth}{!}{%
\begin{tabular}{@{}l|c|ccccccccc|c@{}}
\toprule
Method                       & Reference & \multicolumn{1}{l}{\begin{tabular}[c]{@{}l@{}}pedes-\\ trian\end{tabular}} & rider         & car           & truck         & bus           & mcycle        & bicycle       & \multicolumn{1}{l}{\begin{tabular}[c]{@{}l@{}}traffic \\ light\end{tabular}} & \multicolumn{1}{l|}{\begin{tabular}[c]{@{}l@{}}traffic \\ sign\end{tabular}} & \multicolumn{1}{l}{mAP} \\ \midrule
Source                      & -         & 50.0                                                                       & 28.9          & 66.6          & 47.8          & 47.5          & 32.8          & 39.5          & 41.0                                                                         & 56.5                                                                         & 41.1                    \\
Oracle                      & -         & 52.1                                                                       & 35.0          & 73.6          & 53.5          & 54.8          & 36.0          & 41.8          & 52.2                                                                         & 63.3                                                                         & 46.2                    \\ \midrule
DA-Faster \dag\cite{dafasterrcnn} & CVPR'18   & 50.4                                                                       & 30.3          & 66.3          & 46.8          & 48.3          & 32.6          & 41.4          & 41.0                                                                         & 56.2                                                                         & 41.3                    \\
UMT \dag\cite{umt}               & CVPR'21   & 46.5                                                                       & 26.1          & 46.8          & 44.0          & 46.3          & 28.2          & 40.2          & 31.6                                                                         & 52.7                                                                         & 36.2                    \\
TDD \dag\cite{tdd}                & CVPR'22   & 43.1                                                                       & 20.7          & 68.4          & 33.3          & 35.6          & 16.5          & 25.9          & 43.1                                                                         & 59.5                                                                         & 34.6                    \\
AT \dag\cite{at}                  & CVPR'22   & 42.3                                                                       & 30.4          & 60.8          & 48.9          & 52.1          & 34.5          & 42.7          & 29.1                                                                         & 43.9                                                                         & 38.5                    \\
2PCNet\cite{2pc}             & CVPR'23   & 54.4                                                                       & 30.8          & 73.1          & \textbf{53.8} & 55.2          & 37.5          & 44.5          & \textbf{49.4}                                                                & 65.2                                                                         & 46.4                    \\ \midrule
ADT                          & Ours      & \textbf{55.2}                                                              & \textbf{35.5} & \textbf{73.2} & 53.3          & \textbf{55.5} & \textbf{40.5} & \textbf{47.3} & 46.8                                                                         & \textbf{67.0}                                                                & \textbf{47.5}           \\ \bottomrule
\end{tabular}%
}
\dag Results reproduced by 2PCNet\cite{2pc}.
\end{table}

The results of night adaptation are shown in \cref{tab:d2n}. Similar as in fog adaptation, our ADT outperforms the oracle model by 1.3 \%, verifying the effectiveness of our approach for cross domain object detection under various adverse visibility conditions.

When compared with other methods, ADT establishes a new state-of-the-art, surpassing the previous best-performing model by 1.1 \%. As observed in \cite{2pc}, the combination of Mean Teacher with adversarial learning can suppress night-time features, leading to the underperformance of methods like AT compared to the source model. Our method does not suffer from the same limitations, enabling higher performance. It is noteworthy that, unlike 2PCNet designed specifically for night adaptation, leveraging specialized augmentation techniques, our approach uses general augmentation techniques and achieves superior performance. This validates that adversarial defense contributes to enhanced robustness under diverse poor visibility conditions.

\subsection{Ablation Studies}
\label{ssec:exp:ablation}
We further conduct ablation studies on important components in \cref{tab:ablation} and observe the performance gain from each component.

\subsubsection{Adversarial Defense.}
As summarized in \cref{tab:ablation}, the addition of Adversarial Defense demonstrates huge performance improvements compared to the Mean Teacher baseline on both fog (2.7\% AP) and night (1.2\%) adaptation. To highlight the benefits of combining Mean Teacher and Adversarial Defense, we conduct an additional experiment: within each batch, adversarial attack and defense are implemented with a specific probability $P_{\text{attack}}$. The corresponding results are presented in \cref{tab:attack}. Notably, as we increase $P_{\text{attack}}$, there is a significant rise in mAP.
\begin{table}[tb]
\caption{\textbf{Ablation studies.} The last row indicates the Mean Teacher baseline. AD represents Adversarial Defense and ZZ refers to Zoom-in Zoom-out strategy.}
\label{tab:ablation}
\centering
\resizebox{\textwidth}{!}{%
\begin{tabular}{@{}llccc|ccc@{}}
\toprule
\multicolumn{2}{l}{Methods}                  & \multicolumn{3}{c|}{Cityscapes $\rightarrow$ Foggy Cityscapes} & \multicolumn{3}{c}{BDD100K daytime $\rightarrow$ BDD100K night} \\ \midrule
AD         & \multicolumn{1}{l|}{ZZ}         & mAP@[.5:.95]        & mAP@0.5             & mAP@0.75           & mAP@[.5:.95]        & mAP@0.5             & mAP@0.75            \\ \midrule
\checkmark & \multicolumn{1}{l|}{\checkmark} & \textbf{29.6}       & \textbf{54.5}       & \textbf{27.9}      & \textbf{23.5}       & \textbf{47.5}       & \textbf{20.2}       \\
\checkmark & \multicolumn{1}{l|}{}           & 28.3                & 52.0                & 26.5               & 23.2                & 47.2                & 19.5                \\
           & \multicolumn{1}{l|}{}           & 26.4                & 49.3                & 24.5               & 22.7                & 46.0                & 19.5                \\ \bottomrule
\end{tabular}%
}
\end{table}

\begin{table}[tb]
\caption{Effect of adversarial defense on foggy adaptation given different attack probabilities. $P_{\text{attack}}=0$ represents the Mean Teacher baseline.}
\label{tab:attack}
\centering
\begin{tabular}{@{}c|c|c|c@{}}
\toprule
$P_{\text{attack}}$& mAP@{[}.5:.95{]} & mAP@0.5       & mAP@0.75      \\ \midrule
0           & 26.4             & 49.3          & 24.5          \\
0.3         & 27.3             & 50.3          & 25.3          \\
0.5         & 27.6             & 51.1          & 25.9          \\
0.7         & 27.6             & 51.2          & 26.4          \\
1.0         & \textbf{28.3}    & \textbf{52.0} & \textbf{26.5} \\ \bottomrule
\end{tabular}
\end{table}

\subsubsection{Zoom-in Zoom-out Strategy.} Results in \cref{tab:ablation} clearly show the advantage of the Zoom-in Zoom-out strategy, leading to performance improvements of 2.5\% in foggy adaptation and 0.3\% in night adaptation. The effectiveness of the Zoom-in Zoom-out strategy is observed to be slightly diminished in night adaptation, due to the distinct domain shifts encountered in foggy weather and at night. In nighttime conditions, streetlights and vehicle indicators exhibit visual similarities to traffic lights but on a smaller scale, resulting in less performance improvement.

\section{Conclusions}
\label{sec:conclusions}
In this work, we tackle the challenge of Cross-Domain Object Detection under poor visibility conditions and propose a novel framework \textit{Adversarial Defense Teacher}. We reveal that the limited inconsistency between predictions on weakly and strongly augmented target data hinders the Mean Teacher framework from better teacher-student mutual learning. The integration of adversarial defense into ADT strategically guides the student model to update itself in the most informative direction. This update encourages the model to generalize on subtly perturbed inputs that effectively deceive the model, fostering a more robust adaptation. Additionally, we present a Zoom-in Zoom-out strategy to address small object detection under adverse visibility conditions. This strategy involves zooming in on target images for pseudo-label generation by the teacher model and subsequently zooming out, along with pseudo-labels, for input into the student model. This process compels the student model to detect downscaled objects, refining the learned features. Our experiments validate the effectiveness of the proposed approach, particularly in the context of fog and night adaptations.
\bibliographystyle{splncs04}
\bibliography{egbib}

\begin{thebibliography}{10}
\providecommand{\url}[1]{\texttt{#1}}
\providecommand{\urlprefix}{URL }
\providecommand{\doi}[1]{https://doi.org/#1}

\bibitem{multi_modal2}
Bijelic, M., Gruber, T., Mannan, F., Kraus, F., Ritter, W., Dietmayer, K., Heide, F.: Seeing through fog without seeing fog: Deep multimodal sensor fusion in unseen adverse weather. In: Proceedings of the IEEE/CVF Conference on Computer Vision and Pattern Recognition. pp. 11682--11692 (2020)

\bibitem{translation1}
Bousmalis, K., Silberman, N., Dohan, D., Erhan, D., Krishnan, D.: Unsupervised pixel-level domain adaptation with generative adversarial networks. In: Proceedings of the IEEE conference on computer vision and pattern recognition. pp. 3722--3731 (2017)

\bibitem{cascadercnn}
Cai, Z., Vasconcelos, N.: Cascade r-cnn: Delving into high quality object detection. In: Proceedings of the IEEE conference on computer vision and pattern recognition. pp. 6154--6162 (2018)

\bibitem{cmt}
Cao, S., Joshi, D., Gui, L.Y., Wang, Y.X.: Contrastive mean teacher for domain adaptive object detectors. In: Proceedings of the IEEE/CVF Conference on Computer Vision and Pattern Recognition. pp. 23839--23848 (2023)

\bibitem{argoverse}
Chang, M.F., Lambert, J., Sangkloy, P., Singh, J., Bak, S., Hartnett, A., Wang, D., Carr, P., Lucey, S., Ramanan, D., Hays, J.: Argoverse: 3d tracking and forecasting with rich maps. In: Proceedings of the IEEE/CVF Conference on Computer Vision and Pattern Recognition (CVPR) (June 2019)

\bibitem{htcn}
Chen, C., Zheng, Z., Ding, X., Huang, Y., Dou, Q.: Harmonizing transferability and discriminability for adapting object detectors. In: Proceedings of the IEEE/CVF Conference on Computer Vision and Pattern Recognition. pp. 8869--8878 (2020)

\bibitem{pt}
Chen, M., Chen, W., Yang, S., Song, J., Wang, X., Zhang, L., Yan, Y., Qi, D., Zhuang, Y., Xie, D., et~al.: Learning domain adaptive object detection with probabilistic teacher. In: International Conference on Machine Learning. pp. 3040--3055. PMLR (2022)

\bibitem{dafasterrcnn}
Chen, Y., Li, W., Sakaridis, C., Dai, D., Van~Gool, L.: Domain adaptive faster r-cnn for object detection in the wild. In: Proceedings of the IEEE conference on computer vision and pattern recognition. pp. 3339--3348 (2018)

\bibitem{cityscapes}
Cordts, M., Omran, M., Ramos, S., Rehfeld, T., Enzweiler, M., Benenson, R., Franke, U., Roth, S., Schiele, B.: The cityscapes dataset for semantic urban scene understanding. In: Proceedings of the IEEE Conference on Computer Vision and Pattern Recognition (CVPR) (June 2016)

\bibitem{imagenet}
Deng, J., Dong, W., Socher, R., Li, L.J., Li, K., Fei-Fei, L.: Imagenet: A large-scale hierarchical image database. In: 2009 IEEE conference on computer vision and pattern recognition. pp. 248--255. Ieee (2009)

\bibitem{umt}
Deng, J., Li, W., Chen, Y., Duan, L.: Unbiased mean teacher for cross-domain object detection. In: Proceedings of the IEEE/CVF Conference on Computer Vision and Pattern Recognition. pp. 4091--4101 (2021)

\bibitem{ht}
Deng, J., Xu, D., Li, W., Duan, L.: Harmonious teacher for cross-domain object detection. In: Proceedings of the IEEE/CVF Conference on Computer Vision and Pattern Recognition. pp. 23829--23838 (2023)

\bibitem{kitti}
Geiger, A., Lenz, P., Urtasun, R.: Are we ready for autonomous driving? the kitti vision benchmark suite. In: 2012 IEEE conference on computer vision and pattern recognition. pp. 3354--3361. IEEE (2012)

\bibitem{fasterrcnn}
Girshick, R.: Fast r-cnn. In: Proceedings of the IEEE international conference on computer vision. pp. 1440--1448 (2015)

\bibitem{fgsm}
Goodfellow, I.J., Shlens, J., Szegedy, C.: Explaining and harnessing adversarial examples. arXiv preprint arXiv:1412.6572  (2014)

\bibitem{haze_removal1}
He, K., Sun, J., Tang, X.: Single image haze removal using dark channel prior. IEEE transactions on pattern analysis and machine intelligence  \textbf{33}(12),  2341--2353 (2010)

\bibitem{resnet}
He, K., Zhang, X., Ren, S., Sun, J.: Deep residual learning for image recognition. In: Proceedings of the IEEE conference on computer vision and pattern recognition. pp. 770--778 (2016)

\bibitem{tdd}
He, M., Wang, Y., Wu, J., Wang, Y., Li, H., Li, B., Gan, W., Wu, W., Qiao, Y.: Cross domain object detection by target-perceived dual branch distillation. In: Proceedings of the IEEE/CVF Conference on Computer Vision and Pattern Recognition. pp. 9570--9580 (2022)

\bibitem{maf}
He, Z., Zhang, L.: Multi-adversarial faster-rcnn for unrestricted object detection. In: Proceedings of the IEEE/CVF International Conference on Computer Vision. pp. 6668--6677 (2019)

\bibitem{translation3}
Hsu, H.K., Yao, C.H., Tsai, Y.H., Hung, W.C., Tseng, H.Y., Singh, M., Yang, M.H.: Progressive domain adaptation for object detection. In: Proceedings of the IEEE/CVF winter conference on applications of computer vision. pp. 749--757 (2020)

\bibitem{pda}
Hsu, H.K., Yao, C.H., Tsai, Y.H., Hung, W.C., Tseng, H.Y., Singh, M., Yang, M.H.: Progressive domain adaptation for object detection. In: Proceedings of the IEEE/CVF winter conference on applications of computer vision. pp. 749--757 (2020)

\bibitem{apollo}
Huang, X., Cheng, X., Geng, Q., Cao, B., Zhou, D., Wang, P., Lin, Y., Yang, R.: The apolloscape dataset for autonomous driving. In: Proceedings of the IEEE conference on computer vision and pattern recognition workshops. pp. 954--960 (2018)

\bibitem{2pc}
Kennerley, M., Wang, J.G., Veeravalli, B., Tan, R.T.: 2pcnet: Two-phase consistency training for day-to-night unsupervised domain adaptive object detection. In: Proceedings of the IEEE/CVF Conference on Computer Vision and Pattern Recognition. pp. 11484--11493 (2023)

\bibitem{translation2}
Kim, T., Jeong, M., Kim, S., Choi, S., Kim, C.: Diversify and match: A domain adaptive representation learning paradigm for object detection. In: Proceedings of the IEEE/CVF Conference on Computer Vision and Pattern Recognition. pp. 12456--12465 (2019)

\bibitem{at}
Li, Y.J., Dai, X., Ma, C.Y., Liu, Y.C., Chen, K., Wu, B., He, Z., Kitani, K., Vajda, P.: Cross-domain adaptive teacher for object detection. In: Proceedings of the IEEE/CVF Conference on Computer Vision and Pattern Recognition. pp. 7581--7590 (2022)

\bibitem{retinanet}
Lin, T.Y., Goyal, P., Girshick, R., He, K., Doll{\'a}r, P.: Focal loss for dense object detection. In: Proceedings of the IEEE international conference on computer vision. pp. 2980--2988 (2017)

\bibitem{ssd}
Liu, W., Anguelov, D., Erhan, D., Szegedy, C., Reed, S., Fu, C.Y., Berg, A.C.: Ssd: Single shot multibox detector. In: Computer Vision--ECCV 2016: 14th European Conference, Amsterdam, The Netherlands, October 11--14, 2016, Proceedings, Part I 14. pp. 21--37. Springer (2016)

\bibitem{pgd}
Madry, A., Makelov, A., Schmidt, L., Tsipras, D., Vladu, A.: Towards deep learning models resistant to adversarial attacks. arXiv preprint arXiv:1706.06083  (2017)

\bibitem{raindrop_removal}
Qian, R., Tan, R.T., Yang, W., Su, J., Liu, J.: Attentive generative adversarial network for raindrop removal from a single image. In: Proceedings of the IEEE conference on computer vision and pattern recognition. pp. 2482--2491 (2018)

\bibitem{yolo}
Redmon, J., Divvala, S., Girshick, R., Farhadi, A.: You only look once: Unified, real-time object detection. In: Proceedings of the IEEE conference on computer vision and pattern recognition. pp. 779--788 (2016)

\bibitem{yolo9000}
Redmon, J., Farhadi, A.: Yolo9000: better, faster, stronger. In: Proceedings of the IEEE conference on computer vision and pattern recognition. pp. 7263--7271 (2017)

\bibitem{yolov3}
Redmon, J., Farhadi, A.: Yolov3: An incremental improvement. arXiv preprint arXiv:1804.02767  (2018)

\bibitem{haze_removal2}
Ren, W., Liu, S., Zhang, H., Pan, J., Cao, X., Yang, M.H.: Single image dehazing via multi-scale convolutional neural networks. In: Computer Vision--ECCV 2016: 14th European Conference, Amsterdam, The Netherlands, October 11-14, 2016, Proceedings, Part II 14. pp. 154--169. Springer (2016)

\bibitem{adverse}
Rothmeier, T., Wachtel, D., von Dem Bussche-H{\"u}nnefeld, T., Huber, W.: I had a bad day: Challenges of object detection in bad visibility conditions. In: 2023 IEEE Intelligent Vehicles Symposium (IV). pp.~1--6. IEEE (2023)

\bibitem{sw}
Saito, K., Ushiku, Y., Harada, T., Saenko, K.: Strong-weak distribution alignment for adaptive object detection. In: Proceedings of the IEEE/CVF Conference on Computer Vision and Pattern Recognition. pp. 6956--6965 (2019)

\bibitem{foggy_cityscapes}
Sakaridis, C., Dai, D., Van~Gool, L.: Semantic foggy scene understanding with synthetic data. International Journal of Computer Vision  \textbf{126},  973--992 (2018)

\bibitem{vgg}
Simonyan, K., Zisserman, A.: Very deep convolutional networks for large-scale image recognition. arXiv preprint arXiv:1409.1556  (2014)

\bibitem{mt}
Tarvainen, A., Valpola, H.: Mean teachers are better role models: Weight-averaged consistency targets improve semi-supervised deep learning results. Advances in neural information processing systems  \textbf{30} (2017)

\bibitem{fcos}
Tian, Z., Shen, C., Chen, H., He, T.: Fcos: Fully convolutional one-stage object detection. In: Proceedings of the IEEE/CVF international conference on computer vision. pp. 9627--9636 (2019)

\bibitem{detectron2}
Wu, Y., Kirillov, A., Massa, F., Lo, W.Y., Girshick, R.: Detectron2. \url{https://github.com/facebookresearch/detectron2} (2019)

\bibitem{crda}
Xu, C.D., Zhao, X.R., Jin, X., Wei, X.S.: Exploring categorical regularization for domain adaptive object detection. In: Proceedings of the IEEE/CVF Conference on Computer Vision and Pattern Recognition. pp. 11724--11733 (2020)

\bibitem{bddv}
Xu, H., Gao, Y., Yu, F., Darrell, T.: End-to-end learning of driving models from large-scale video datasets. In: Proceedings of the IEEE conference on computer vision and pattern recognition. pp. 2174--2182 (2017)

\bibitem{rainstreak_removal}
Yang, W., Tan, R.T., Feng, J., Liu, J., Guo, Z., Yan, S.: Deep joint rain detection and removal from a single image. In: Proceedings of the IEEE conference on computer vision and pattern recognition. pp. 1357--1366 (2017)

\bibitem{bdd}
Yu, F., Chen, H., Wang, X., Xian, W., Chen, Y., Liu, F., Madhavan, V., Darrell, T.: Bdd100k: A diverse driving dataset for heterogeneous multitask learning. In: Proceedings of the IEEE/CVF conference on computer vision and pattern recognition. pp. 2636--2645 (2020)

\bibitem{multi}
Zhou, W., Du, D., Zhang, L., Luo, T., Wu, Y.: Multi-granularity alignment domain adaptation for object detection. In: Proceedings of the IEEE/CVF Conference on Computer Vision and Pattern Recognition. pp. 9581--9590 (2022)

\bibitem{selective}
Zhu, X., Pang, J., Yang, C., Shi, J., Lin, D.: Adapting object detectors via selective cross-domain alignment. In: Proceedings of the IEEE/CVF Conference on Computer Vision and Pattern Recognition. pp. 687--696 (2019)

\end{thebibliography}

%
\pagebreak
\appendix
\section{Pseudo-code for Adversarial Defense Teacher}
In \cref{alg:adt}, we present the pseudo-code of our Adversarial Defense Teacher (ADT) framework. The key difference between ADT and the traditional Mean Teacher (MT) framework is \textcolor{blue}{highlighted}.
\begin{algorithm}
\caption{Adversarial Defense Teacher}\label{alg:adt}
\textbf{Input: }Labeled source data $\{X_s,B_s,C_s\}$ and unlabeled target data $\{X_t\}$. \\
\textbf{Output: }Student $\mathcal{D}(\cdot,\theta_{\text{student}})$ and Teacher $\mathcal{D}(\cdot,\theta_{\text{teacher}})$ after domain adaptation
\begin{algorithmic}[1]
\State Pretrain a source model $\mathcal{D}(\cdot,\theta_{\text{source}})$ on $\{X_s,B_s,C_s\}$.
\State $\theta_{\text{student}} \gets \theta_{\text{source}},\ \theta_{\text{teacher}} \gets \theta_{\text{source}},\ \text{iteration} \gets 0$
\While{$\text{iteration} \neq T_{\text{max\_iteration}}$}
    \State Get batch of labeled source data $\{x_s,b_s,c_s\}$ and unlabeled target data $\{x_t\}$
    \State Conduct strong augmentation on $x_s$ to generate $x_s^s$
    \State Conduct weak and strong augmentation on $x_t$ to generate $x_t^w, x_t^s$
    \textcolor{blue}{
    \State $\texttt{// Zoom-in}$
    \State Upscale $x_t^w$ with a randomly selected zoom-in ratio $r_{\text{in}}>1$: \newline\centerline{$x_t^{\text{in}}=\text{Scale}(x_t^w,r_{\text{in}} )$}}
    \State Update Teacher by EMA as in Eq. (5): $\theta_{\text{teacher}}\gets \beta\theta_{\text{teacher}}+(1-\beta)\theta_{\text{student}}$
    \State Teacher generates pseudo-labels on \textcolor{blue}{$x_t^{\text{in}}$}: $b_t,c_t=\text{Filter}(\mathcal{D}(\textcolor{blue}{x_t^{\text{in}}},\theta_{\text{teacher}}),\tau)$
    \textcolor{blue}{
    \State $\texttt{// Zoom-out}$
    \State Downscale $x_t^s$ with a randomly selected zoom-out ratio $r_{\text{out}}<1$:\newline\centerline{$x_t^{\text{out}}=\text{Scale}(x_t^s,r_{\text{out}})$}
    \State Resize pseudo-labels for $x_t^{\text{out}}$:\ $b_t',c_t'=\text{Resize}(b_t,c_t,r_{\text{in}},r_{\text{out}})$
    \State $\texttt{// Adversarial Defense}$
    \State $i \gets 0,\ x_{\text{adv}}^i\gets x_t^{\text{out}}$
    \While{$i \neq T_{\text{max\_attack}}$}
        \State Compute $\mathcal{L}_{\text{attack}}^i$ according to Eq. (10):\newline \centerline{$\mathcal{L}_{\text{attack}}^i=\mathcal{L}^{\text{rpn}}_{\text{attack}}(x_{\text{adv}}^i,b_t',c_t')+\mathcal{L}^{\text{roi}}_{\text{attack}}(x_{\text{adv}}^i,b_t',c_t')$}
        \State Compute $x_{\text{adv}}^{i+1}$ according to Eq. (7):\newline \centerline{$x_{\text{adv}}^{i+1}=\text{Clip}_{x_{\text{adv}}^{0},\epsilon}\left(x_{\text{adv}}^{i}+\alpha\cdot \text{sgn}(\nabla_x \mathcal{L}_{\text{attack}}^i)\right)$}
        \State $i\gets i+1$
    \EndWhile
    \State $x_{\text{adv}} \gets x_{\text{adv}}^{T_{\text{max\_attack}}}$}
    \State Compute source loss as in Eq. (1): $\mathcal{L}_{s} = \mathcal{L}^{\text{rpn}}(x_s^s,b_s,c_s)+\mathcal{L}^{\text{roi}}(x_s^s,b_s,c_s)$
    \State Compute target loss as in Eq. (2): $\mathcal{L}_t=\mathcal{L}^{\text{rpn}}(\textcolor{blue}{x_{\text{adv}}},b_t',c_t')+\mathcal{L}^{\text{roi}}(\textcolor{blue}{x_{\text{adv}}},b_t',c_t')$
    \State Compute total loss: $\mathcal{L}=\lambda_s\mathcal{L}_{s} + \lambda_t\mathcal{L}_{t}$
    \State Update Student: $\theta_{\text{student}}=\theta_{\text{student}}-\eta\nabla_\theta\mathcal{L}$
    \State $\text{iteration} \gets \text{iteration}+1$
\EndWhile
\end{algorithmic}
\end{algorithm}

\section{Qualitative Results}
\label{sec:qr}
\begin{figure}[tb]
  \centering
  \includegraphics[width=\textwidth]{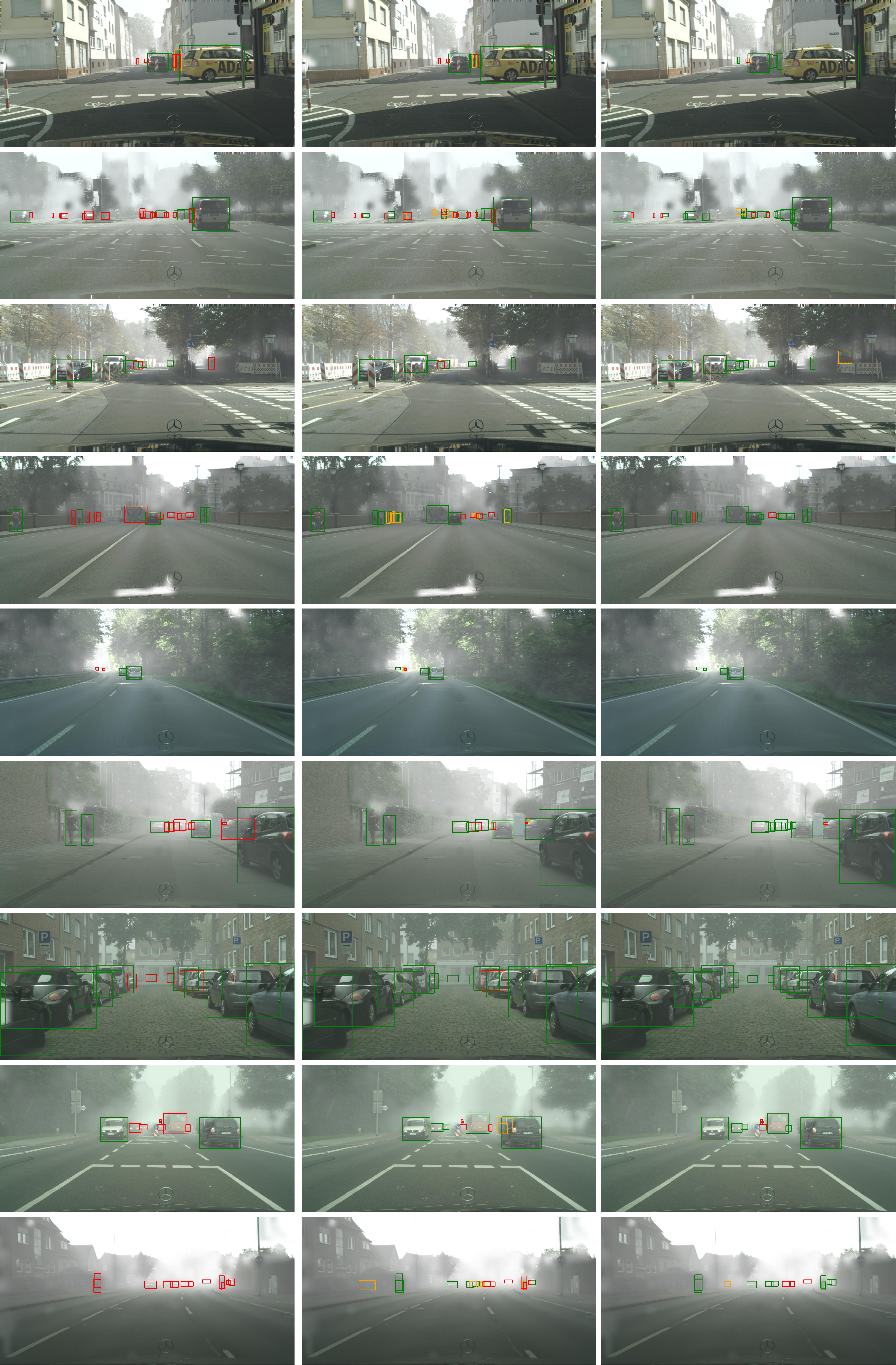}
  \caption{Qualitative results of foggy adaptation for source model (left column), CMT (middle column) and Ours (right column). \textcolor{ForestGreen}{Green}, \textcolor{red}{red} and \textcolor{orange}{orange} boxes denote true positives, false negatives and false positives, respectively. We set the score threshold to 0.8 and evaluate all models on images resized to a shorter side of 600 pixels.}
  \label{fig:qr}
\end{figure}

We present qualitative results comparing the state-of-the-art (SOTA) method, Contrastive Mean Teacher (CMT), with our ADT on the Cityscapes $\rightarrow$ Foggy Cityscapes benchmark in \cref{fig:qr}.
\end{document}